\documentclass{article}
\usepackage{float}
\usepackage{spconf,amsmath,graphicx}
\usepackage{amssymb}
\usepackage{amsfonts}
\usepackage{algorithm}
\usepackage{algorithmic}
\usepackage{booktabs}
\usepackage{xcolor}
\usepackage{mathrsfs}
\usepackage{hyperref}
\usepackage{cleveref} 
\usepackage{url}
\usepackage{nicefrac}
\usepackage{microtype}
\usepackage{mathtools}

\title{Adaptive Multimodal Protein Plug-and-Play with Diffusion-Based Priors}
\name{Amartya Banerjee$^*$, Xingyu Xu$^\dagger$, Caroline Moosmüller$^*$, Harlin Lee$^*$ \thanks{This work was partially supported by the AI Acceleration Program at the University of North Carolina at Chapel Hill and NSF DMS-2410140. The authors thank Yuejie Chi for initiating this project in the summer of 2024. Contact: amartya1@cs.unc.edu.}}
\address{University of North Carolina at Chapel Hill$^*$, Carnegie Mellon University$^\dagger$}

\begin{document}
%
\maketitle
\begin{abstract}
In an inverse problem, the goal is to recover an unknown parameter (e.g., an image) that has typically undergone some lossy or noisy transformation during measurement. Recently, deep generative models, particularly diffusion models, have emerged as powerful priors for protein structure generation. However, integrating noisy experimental data from multiple sources to guide these models remains a significant challenge. Existing methods often require precise knowledge of experimental noise levels and manually tuned weights for each data modality. In this work, we introduce Adam-PnP, a Plug-and-Play framework that guides a pre-trained protein diffusion model using gradients from multiple, heterogeneous experimental sources. Our framework features an adaptive noise estimation scheme and a dynamic modality weighting mechanism integrated into the diffusion process, which reduce the need for manual hyperparameter tuning.  Experiments on complex reconstruction tasks demonstrate significantly improved accuracy using Adam-PnP.
\end{abstract}
\begin{keywords}
Diffusion models, inverse problems, protein backbone reconstruction, multimodal learning
\end{keywords}
\section{Introduction}
\label{sec:intro}

The determination of three-dimensional protein structures is fundamental for understanding biological function and for novel drug design. While experimental techniques like X-ray crystallography \cite{drenth2007principles}, cryo-electron microscopy (cryo-EM) \cite{zhang2003visualization}, and nuclear magnetic resonance (NMR) spectroscopy \cite{wuthrich1989protein} provide invaluable structural information, they each have limitations. Data from these methods can be sparse, noisy, ambiguous, and often incomplete. Computational methods are crucial for integrating these diverse data streams to produce high-resolution structural models.
Recently, deep generative models, especially denoising diffusion probabilistic models (DDPMs) \cite{ho2020denoising}, have demonstrated unprecedented success in generating realistic and designable protein structures \cite{chroma_ingraham2023illuminating}.
These models learn a rich, implicit prior over the space of valid protein conformations \cite{watson2023novo, geffner2025laproteinaatomisticproteingeneration}. Recent efforts have focused on applying these powerful priors to challenging inverse problems like cryo-EM map interpretation and de novo protein design \cite{maddipatlainverse, advaith2024generative,levy2024solving}.

In this work, we introduce Adam-PnP, a Plug-and-Play framework that guides a pre-trained protein diffusion model using gradients from multiple, heterogeneous experimental sources. Our key contributions are twofold: (1) an adaptive noise estimation scheme that learns the unknown noise level of each modality on-the-fly, and (2) a dynamic modality weighting mechanism that automatically balances the influence of each data source based on its estimated reliability. We demonstrate our method's effectiveness on complex reconstruction tasks, showing that combining modalities improves accuracy, with a combination of partial atomic coordinates, distance restraints, and cryo-EM density maps. Our experiments on a multimodal reconstruction task show that combining modalities significantly improves accuracy, with a combination of partial coordinates and distance restraints achieving a state-of-the-art backbone RMSD of $0.65 \pm 0.18$ \AA. Thanks to its adaptive design, our method remains versatile across diverse scenarios, promising advantageous practical utility. Code will be available at \href{https://github.com/amartya21/Adam-PnP}{\url{https://github.com/amartya21/Adam-PnP}}.

\section{Preliminaries}
\subsection{Problem Formulation}
We represent a protein backbone by its 3-dimensional coordinates $\mathbf{x} \in \mathbb{R}^{4N \times 3}$, where $N$ is the number of amino acid residues. For $i=1,\ldots, M$, we observe noisy measurements
\begin{equation}
y_i = \mathcal{F}_i(\mathbf{x}^*) + \eta_i,~\eta_i \sim \mathcal{N}(\mathbf{0}; \sigma_{i}^2 \mathbf{I}), \label{eq:data_model}
\end{equation}
where noise level $\sigma_{i}$ is unknown, but the forward model $\mathcal{F}_i$ is known and differentiable. $\eta_i$ is independent of $\mathcal{F}_i$. The dimensions of $y_i$ and $\eta_i$ may be different for different $\mathcal{F}_i$. In this multimodal inverse problem, we aim to estimate ground truth $\mathbf{x}^*$ given $\{\mathcal{F}_i\}$ and $\mathbf{Y} := \{y_i\}$.

\subsection{Diffusion Models for Protein Structure Generation}
Diffusion models learn to reverse a forward process that gradually adds noise to data. Let $\mathbf{x}_0$ be the initial data at $t=0$.  
The forward process is a stochastic differential equation (SDE) that evolves $\mathbf{x}_0$ to a simple noise distribution at time $t=T$ via
\begin{equation}
    d\mathbf{x}_t = f(t) \mathbf{x}_t \mathrm dt + g(t) \mathrm d\mathbf{W}_t, \label{eq:forward_sde}
\end{equation}
where $\mathbf{W}_t$ is a standard Wiener process \cite{songscore} and $f,g: [0,T] \to \mathbb{R}$ are deterministic functions. The generative process is the corresponding reverse-time SDE:
\begin{equation}
    d\mathbf{x}_t = \left[ f(t)\mathbf{x}_t - g(t)^2 \mathbf{s}_{\theta}(\mathbf{x}_t, t) \right] \mathrm dt + g(t) \mathrm d\overline{\mathbf{W}}_t. \label{eq:reverse_sde}
\end{equation}
Here $\overline{\mathbf{W}}$ is a reverse-time Wiener process, and $\mathbf{s}_{\theta}(\mathbf{x}_t, t) \approx \nabla_{\mathbf{x}_t} \log p_t(\mathbf{x}_t)$ is the score function, which can be approximated using a trained denoising neural network $D_\theta(\mathbf{x}_t, t)$.

Though any pretrained diffusion model can be used as $D_\theta$, in our work, this is handled by the Chroma protein backbone generation model~\cite{chroma_ingraham2023illuminating}.

To preserve a protein's fundamental biophysical properties, Chroma employs a non-isotropic noising process governed by a covariance matrix $\mathbf{R} \in \mathbb{R}^{4N \times 4N}$.
The forward SDE is 
\begin{equation}
d\mathbf{x}_t = -\frac{\beta_t}{2}\mathbf{x}_t \mathrm dt + \sqrt{\beta_t}\mathbf{R} \, \mathrm d\mathbf{W}_t,
\label{eq:forward_chroma_sde}
\end{equation}
 where $\beta_t$ is a deterministic function.

For generation, we use the hybrid Langevin reverse-time SDE introduced in \cite{chroma_ingraham2023illuminating} as an extension of \eqref{eq:reverse_sde}:
\begin{equation}
\begin{split}
d\mathbf{x}_t = {}&
  \Bigl(
    -\tfrac{1}{2}\mathbf{x}_t
    -\bigl(\lambda_t + \tfrac{\lambda_0\psi}{2}\bigr)
      \mathbf{R}\mathbf{R}^{\mathsf{T}}
      \mathbf{s}_{\theta}(\mathbf{x}_t, t)
  \Bigr) \beta_t \, \mathrm dt \\
& \quad + \, \sqrt{\beta_t(1 + \psi)} \, \mathbf{R} \, \mathrm d \overline{\mathbf{W}}_t.
\end{split}
\label{eq:reverse_sde_chroma}
\end{equation}
where $\lambda_t$ is a time-dependent temperature annealing schedule, and $\lambda_0, \psi$ control the final inverse temperature and the rate of Langevin equilibration, respectively.

It is more numerically stable to work in a ``whitened" latent space defined by the transformation $\mathbf{z} = \mathbf{R}^{-1}\mathbf{x}$.

\subsection{Plug-and-Play (PnP) for MAP Estimation}
An inverse problem aims to recover a structure $\mathbf{x}$ from measurements $\mathbf{Y}=\{y_i\}$. In a Bayesian setting, the solution is characterized by the posterior distribution $p(\mathbf{x}|\mathbf{Y}) \propto p(\mathbf{Y}|\mathbf{x})p(\mathbf{x})$, which combines the experimental likelihood $p(\mathbf{Y}|\mathbf{x})$ with the generative prior $p(\mathbf{x})$.
In many structural biology workflows, the primary objective is to determine the single, highest-quality atomic model that best explains the experimental data. This goal aligns with \textit{Maximum A Posteriori} (MAP) estimation, which seeks to find the mode of the posterior distribution:
\begin{align}
  \hat{\mathbf{x}}
    &= \arg\max_{\mathbf{x}} p(\mathbf{x}\mid\mathbf{Y})  \nonumber \\
    &= \arg\min_{\mathbf{x}}
      \left\{ -\sum_{i=1}^M \log p(y_i\mid \mathbf{x}) - \log p(\mathbf{x})\right\}\label{eq:map}
\end{align}
Recall that the prior $p(\mathbf{x})$ can be approximated using pretrained denoiser $D_\theta$.
Leveraging this idea, the Plug-and-Play (PnP) framework \cite{venkatakrishnan2013plug, graikos2022diffusion, xu2024provably} optimizes \eqref{eq:map} by alternating between a likelihood-enforcement step (multimodal data consistency) and a prior-enforcement step (structural realism). 

\section{Methodology}
\label{sec:methodology}
\Cref{alg:adam-pnp} implements this PnP strategy as an iterative refinement process within the diffusion model's reverse SDE in \eqref{eq:reverse_sde_chroma}. It uniquely integrates multimodal likelihoods with a novel adaptive engine for noise estimation and weighting. We emphasize that Adam-PnP is modular and can accommodate any differentiable forward model $\mathcal{F}_i(\mathbf{x})$ that maps a structure to an experimental observable.

\subsection{Iterative PnP Updates}
\label{sec:likelihoods}
Within each reverse diffusion timestep $t$, we perform two updates corresponding to the two terms in \eqref{eq:map}. First, we use only the prior model $p(\mathbf{x})$; that is, we use the denoiser $D_\theta$ to obtain an estimate of the clean structure, $\tilde{\mathbf{z}}_0 = \mathbf{R}^{-1} D_\theta(\mathbf{R}\mathbf{z}_t, t)$ in the latent space. Next, we perform a gradient descent step towards the direction that increases consistency with the measurements $\mathbf{Y}$. Given \eqref{eq:data_model}, the log-likelihood for each modality $y_i$ is:
\begin{equation}
\log p(y_i | \mathbf{R}\mathbf{z}; \sigma_{i}) = -\frac{1}{2\sigma_{i}^2} \|y_i - \mathcal{F}_i(\mathbf{R}\mathbf{z})\|^2 + C, \label{eq:log_likelihood}
\end{equation}
from which we can calculate its derivative. 

\subsection{Adaptive Noise Estimation and Dynamic Weighting}
\label{sec:adaptive_noise}
A key challenge in PNP methods is that the measurement noise variance $\sigma^2_i$ for each modality $i$ is often unknown. A natural approach is to estimate this variance online using the current denoised estimate of the structure, $\tilde{\mathbf{z}}_0 = \mathbf{R}^{-1}D_\theta(\mathbf{R}\mathbf{z}_t, t)$. However, a naive estimator based on the squared residuals is systematically biased, because the denoiser's prediction $\tilde{\mathbf{z}}_0$ is itself an imperfect estimate of the true latent structure $\mathbf{z}^*$.

To address this, we introduce a novel online method to estimate the noise variance $\sigma^2_{i}$ given $\tilde{\mathbf{z}}_0$. 
Consider the residual for modality $i$, $r_i = y_i - \mathcal{F}_i(\mathbf{R}\tilde{\mathbf{z}}_0)$. By substituting \eqref{eq:data_model}, we can decompose the residual into:

\begin{equation*}
    r_i = \eta_i - \underbrace{\left( \mathcal{F}_i(\mathbf{R}\tilde{\mathbf{z}}_0) - \mathcal{F}_i(\mathbf{R}\mathbf{z}^*) \right)}_{\text{Bias term from denoising error}}.
\end{equation*}
Using $L_i$, the Lipschitz constant of $\mathcal{F}_i$, we can bound: 
\begin{align*}
    \mathbb{E}[\|r_i\|^2] &= \sigma_i^2 + \mathbb{E}[\|\mathcal{F}_i(\mathbf{R}\tilde{\mathbf{z}}_0) - \mathcal{F}_i(\mathbf{R}\mathbf{z}^*)\|^2]\\
   &\le \sigma_i^2+ L_i^2 \mathbb{E}[\|\mathbf{R}\tilde{\mathbf{z}}_0 - \mathbf{R}\mathbf{z}^*\|^2] \le \sigma_i^2+ c_t(L_i \tau_t)^2
\end{align*}
for some constant $c_t$.
This motivates our bias-corrected variance estimator. We first compute an empirical variance from the median of component-wise squared residuals, where the median is taken over all scalars in the residual vector:
\begin{equation}
  \tilde{\sigma}^2_{i} = \mathrm{median}\left( \left[y_i - \mathcal{F}_i(\mathbf{R}\tilde{\mathbf{z}}_0)\right]^2 \right). \label{eq:sigma_tilde}
\end{equation}
We then subtract our approximation of the bias, which is annealed as $t \to 0$ to reflect the denoiser's increasing accuracy:
\begin{equation}
 \hat\sigma^2_{i,t} = \mathrm{EMA}\left[ \max \left(\epsilon, \tilde{\sigma}^2_{i} - \gamma(1 - t)(L_i\,\tau_t)^2 \right)\right].
\label{eq:bias_correction}
\end{equation}
Here, $\gamma, \epsilon > 0$ are hyperparameters, and the final estimate is stabilized across time steps using an exponential moving average (EMA). These online noise estimates give us a way to automatically fuse the multimodal gradients. 

At each time step $t$, we set the weight $w_{i,t}$ for each modality to be inversely proportional to its estimated variance, thereby weighting each gradient by its \textit{precision}:
\begin{equation}
    w_{i,t} \propto \frac{1}{\hat{\sigma}^2_{i,t} + \epsilon}.
    \label{eq:dynamic_weights}
\end{equation}
The weights are normalized such that $\sum_i w_{i,t} = M$, which preserves the overall magnitude of the guidance signal while prioritizing more reliable data sources.

\subsection{Latent Space Update with Momentum}
\label{sec:momentum_update}
The aggregated gradient is used to update $\tilde{\mathbf{z}}_0$. To ensure stable and efficient convergence, we incorporate momentum \cite{polyak1964some}:
\begin{align}
    \mathbf{g}_{i,t} &= \nabla_{\mathbf{z}}\log p(y_i | \mathbf{Rz}; \hat{\sigma}_{i}) \Big|_{\mathbf{z}=\tilde{\mathbf{z}}_0}
    \label{eq:gradient_single_modality}\\
    \mathbf{g}_{\text{total}, t} &= \sum_{i=1}^{M} w_{i,t} \mathbf{g}_{i,t} \label{eq:total_gradient} \\
    \mathbf{v}_{t} &= \rho \mathbf{v}_{t+1} + (1-\rho) \mathbf{g}_{\text{total}, t} \label{eq:momentum_update_v} \\
    \hat{\mathbf{z}}_0 &= \tilde{\mathbf{z}}_0 - \eta \mathbf{v}_{t} \label{eq:guidance_step}
\end{align}
where $\mathbf{v}_t$ is the momentum vector, $\rho$ is the momentum decay parameter, and $\eta$ is the learning rate. This corrected estimate $\hat{\mathbf{z}}_0$ is then used as the starting point for the next reverse diffusion step.

\begin{algorithm}[!ht]
   \small
   \caption{Multimodal Adam-PNP with Diffusion Prior}
   \label{alg:adam-pnp}
\begin{algorithmic}[1]
   \STATE {\bfseries Input:} Measurements $\{y_i\}_{i=1}^M$, forward models $\{\mathcal{F}_i\}$, diffusion denoiser $D_\theta$.
   \STATE Initialize $\mathbf{z}_T \sim \mathcal{N}(0, \mathbf{I})$,  $\mathbf{v}_T = 0$ 
   \FOR{$t = T, ..., 1$}
        \STATE $\tilde{\mathbf{z}}_0 \leftarrow \mathbf{R}^{-1} D_\theta(\mathbf{R}\mathbf{z}_t, t)$ \COMMENT{Denoising step (prior projection)}
        \IF{adaptive estimation is enabled}

            \STATE Estimate noise variance $\hat{\sigma}^2_{i}$ via \eqref{eq:sigma_tilde} and \eqref{eq:bias_correction}.
            \STATE Compute dynamic weights $w_{i,t}$ using \eqref{eq:dynamic_weights} and normalize.
        \ENDIF
        \STATE Compute total weighted gradient $\mathbf{g}_{\text{total}, t}$ using \eqref{eq:total_gradient}.
        \STATE Update moment $\mathbf{v}_{t}$ using \eqref{eq:momentum_update_v}.
        \STATE Apply guidance $\hat{\mathbf{z}}_0 \leftarrow \tilde{\mathbf{z}}_0 - \eta \mathbf{v}_{t}$ using \eqref{eq:guidance_step}.
        \STATE $\mathbf{z}_{t-1} \sim \mathcal{N}(\alpha_{t-1|t}\hat{\mathbf{z}}_0, \tau_{t-1|t}^2 \mathbf{I})$ \COMMENT{Noise injection for next step}
   \ENDFOR
\STATE {\bfseries return} Final estimated structure $\hat{\mathbf{x}} = \mathbf{R}\mathbf{z}_0$.
\end{algorithmic}
\end{algorithm}
\section{Experimental Results}
We evaluated Adam-PnP on the task of reconstructing the 127-residue protein with \texttt{PDB ID: 7r5b} from various combinations of three data modalities \cite{maddipatlainverse,levy2024solving}: 

\begin{itemize}
    \item \textbf{Partial C${\alpha}$ coordinates (P):} $\mathcal{F}_P(\mathbf{x}) = P\mathbf{x}$ is a masking operator that selects a known subset of coordinates.
    \item \textbf{Pairwise C${\alpha}$ distances (D):} $\mathcal{F}_D(\mathbf{x})$ computes a vector of pairwise C$_\alpha$-C$_\alpha$ distances.
    \item \textbf{Simulated low-resolution (2.0 \AA) electron density map (E):} $\mathcal{F}_E(\mathbf{x})$ is a rendering function that simulates a cryo-EM density map from atomic coordinates and computes its low-resolution Fourier coefficients.
\end{itemize}

We designed two sets of experiments to assess: (1) the synergistic effect of fusing different modalities and (2) the framework's performance under data-limited conditions. Values reported are mean $\pm$ standard deviation over three random seeds.

The reconstruction quality is primarily evaluated by the backbone C$\alpha$-Root Mean Square Deviation (RMSD) in Angstroms (\AA) between the lowest-energy sampled structure and the ground-truth crystal structure, after alignment \cite{coutsias2004using,chroma_ingraham2023illuminating}.
\begin{equation*}
    \text{RMSD}(\hat{\mathbf{x}}, \mathbf{x}^*) = \min_{\mathbf{T}\in \mathsf{SO}(3)}\|\mathbf{T } \hat{\mathbf{x}} - \mathbf{x}^*\|.
\end{equation*}

\subsection{Improvement of Multimodal Fusion}
Our first experiment assessed performance using different combinations of modalities, with results averaged across several noise levels ($0.2$ \text{\AA} to $0.7$ \text{\AA} in increments of $0.1$ \text{\AA}  to ensure robust evaluation). The key finding, summarized in Table \ref{tab:results}, is that fusing high-resolution data sources can significantly improve reconstruction accuracy.

\begin{figure*}[ht]
  \centering
  \includegraphics[width=0.655\textwidth]{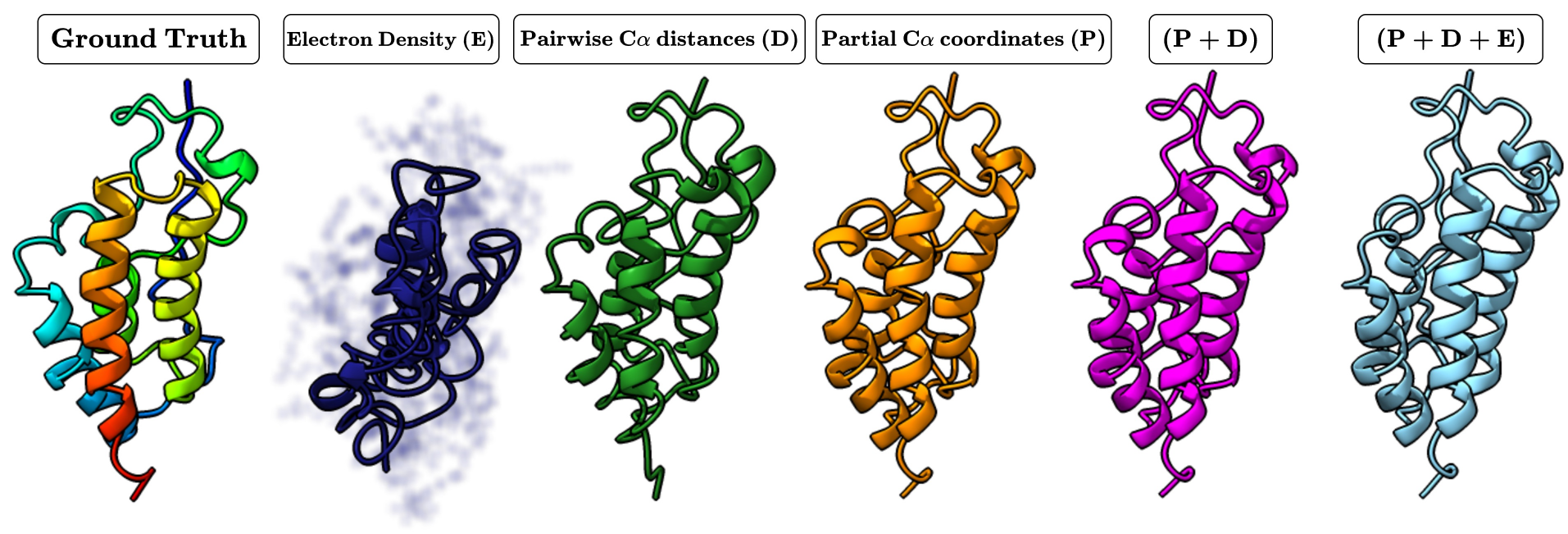}
  \caption{
  Multi-modal protein backbone reconstruction of the protein with \text{PDB ID: 7r5b}. 
  From \textbf{left} to \textbf{right}:  
    \textbf{(GT)} the deposited ground-truth backbone;  
    \textbf{(E)} reconstruction using only the experimental cryo-EM density map (semi-transparent isosurface shown in grey‐blue);  
    \textbf{(D)} reconstruction from pairwise $C_\alpha$ distances;  
    \textbf{(P)} reconstruction from a sparse set of partial $C_\alpha$ coordinates;  
    \textbf{(P\,+\,D)} joint reconstruction from partial coordinates and distances;  
    \textbf{(P\,+\,D\,+\,E)} full multi-modal reconstruction combining all three measurement types. 
  }
  \label{fig:protein_recons}
\end{figure*}

\begin{table}[!h]
\centering
\small
\begin{tabular}{lcc}
\toprule
\textbf{Modalities} & \textbf{Best RMSD (\text{\AA})$\downarrow$} & \textbf{Improvement over P} \\
\midrule
P (Partial Coords) & $0.74 \pm 0.26$ & reference \\
D (Distances)      & $1.17 \pm 0.95$ &  - \\
E (Density)        & $11.67 \pm 1.00$ & - \\
\midrule
\textbf{P+D}              & $\mathbf{0.65 \pm 0.18}$ & $\mathbf{12\%}$\\
P+E                & $0.76 \pm 0.27$ & -$3\%$ \\
D+E                & $1.04 \pm 0.08$ & -$41\%$ \\
\textbf{P+D+E}              & $\mathbf{0.67 \pm 0.19}$ & $\mathbf{9\%}$ \\
\bottomrule
\end{tabular}
\caption{Reconstruction RMSD of Adam-PNP with different modality combinations. 
Bolded rows show that adding D leads to improvement over just P.} \label{tab:results}
\end{table}

Individually, partial coordinates (P) provided the most effective guidance, achieving a sub-Angstrom RMSD. In stark contrast, using only the low-resolution electron density map (E) failed to produce a folded structure, stagnating at an RMSD of 11.98 \text{\AA}. This result highlights the fundamentally ill-posed nature of the problem: the global, low-frequency information in a 2.0 \text{\AA} density map is too ambiguous to guide the model out of the vast landscape of incorrect conformations.

The combination of partial coordinates and distance restraints (P+D) yielded the best overall performance, achieving an exceptional RMSD of $0.65 \pm 0.18$ \text{\AA}. This demonstrates a powerful combination between two distinct but complementary sources of high-resolution spatial information. Interestingly, this P+D combination slightly outperformed the P+D+E combination ($0.67 \pm 0.19$ \text{\AA}). This suggests that the inclusion of the low-resolution `E' modality introduces conflicting, low-quality gradients that can slightly hinder final convergence. However, through the experiment, we can observe that the model learns to dynamically down-weight the influence of the noisy data (using the dynamic weighting scheme in Eq. \ref{eq:dynamic_weights}), effectively trusting the more precise measurements and preventing the low-resolution information from corrupting the final reconstruction. 

\subsection{Ablation with Different Measurements}
To evaluate more realistic scenarios where the full measurement may not be present, we performed an ablation where we varied the number of available observations for the two high-resolution modalities (P and D), while keeping all three (P+D+E) active. The number of partial coordinates supplied ranged from just $22$ ($17\%$ of the structure) to $126$ ($99\%$ of the structure), evaluated at noise levels of $0.05, 0.1,$ and $0.2$ \text{\AA}. The results in Table \ref{tab:ablation_results} show that reconstruction accuracy grows with the amount of high-resolution data.

\begin{table}[ht]
  \footnotesize
  \centering
  \begin{tabular}{@{}ccccc@{}}
    \toprule
    $N_{\text{Distances}}$ &
    $N_{\text{Partial}}$ (\%) &
    \multicolumn{3}{c}{\textbf{Mean C$_{\alpha}$-RMSD (\AA)}} \\
    \cmidrule(lr){3-5}
     & & $\sigma = 0.05$ & $\sigma = 0.10$ & $\sigma = 0.20$ \\
    \midrule
      250 & 22  (17\%) & $2.09\pm0.18$ & $2.18\pm0.13$ & $2.04\pm0.29$ \\
      500 & 32  (25\%) & $1.00\pm0.13$ & $0.98\pm0.13$ & $1.13\pm0.08$ \\
     1000 & 45  (35\%) & $0.75\pm0.06$ & $0.71\pm0.08$ & $0.91\pm0.03$ \\
     2000 & 63  (50\%) & $0.73\pm0.04$ & $0.70\pm0.05$ & $0.92\pm0.04$ \\
     4000 & 89  (70\%) & $\mathbf{0.10\pm0.001}$ & $0.18\pm0.001$ & $0.35\pm0.003$ \\
     8000 & 126 (99\%) & $\mathbf{0.10\pm0.007}$ & $0.18\pm0.003$ & $0.35\pm0.003$ \\
    \bottomrule
  \end{tabular}
  \caption{Reconstruction is dependent on the data sparsity. The quantity of high-resolution data ($N_{\text{Distances}}$ and $N_{\text{Partial}}$) was varied across three noise levels ($\sigma=0.05, 0.1, 0,2$). 
  }
  \label{tab:ablation_results}
\end{table}

\subsubsection{Adaptive Noise Estimation}
 A key feature of our framework is its ability to learn the noise level ($\sigma$) for each modality directly from the data. For the high-resolution distance modality (D), the learned noise parameter, $\hat{\sigma}_{D}$, closely tracked the true underlying noise, $\sigma_{D}$, averaged across all data sparsity conditions; See Table~\ref{tab:noise_estimation}.

 \begin{table}[h]
  \centering
  \small
  \begin{tabular}{@{}ccc@{}}
    \toprule
    \text{True noise $\sigma_{D}$ (\AA)} &
    \text{Estimated $\hat{\sigma}_{D}$ (\AA)} &
    \text{Absolute Error (\AA)} \\
    \midrule
    0.05 & $0.086 \pm 0.009$ & $0.036 \pm 0.009$ \\
    0.10 & $0.120 \pm 0.013$ & $0.020 \pm 0.013$ \\
    0.20 & $0.247 \pm 0.056$ & $0.047 \pm 0.056$ \\
    \bottomrule
  \end{tabular}
  \caption{Ground-truth versus estimated noise for $\mathcal{F}_D$. 
  }
  \label{tab:noise_estimation}
\end{table}
\section{Conclusion}
In this work, we have presented a framework for protein structure determination that effectively integrates a powerful diffusion prior with multiple, heterogeneous experimental data modalities. Our results demonstrate that by fusing complementary high-resolution data, our method can achieve sub-Angstrom accuracy. Furthermore, we have shown that the framework is highly robust to data sparsity, with the generative prior providing essential regularization. The adaptive noise estimation mechanism allows the model to intelligently weigh information from different sources, correctly identifying and down-weighting low-quality data to improve final model accuracy. This work represents a step towards a unified approach for protein structure determination that can flexibly incorporate diverse experimental evidences to solve complex biological inverse problems, particularly for systems where experimental evidence is scarce or of mixed quality.

\newpage
\bibliographystyle{IEEEbib}
\bibliography{proteindiff}

\end{document}